\documentclass[conference]{IEEEtran}
\IEEEoverridecommandlockouts
\usepackage{cite}
\usepackage{amsmath,amssymb,amsfonts}
\usepackage{algorithmic}
\usepackage[ruled,vlined]{algorithm2e}
\usepackage{caption}
\usepackage{listings}
\usepackage[english]{babel}
\usepackage{xcolor}

\usepackage[english]{babel}

\usepackage{subcaption}
\usepackage{graphicx}
\usepackage{float}
\usepackage{textcomp}
\usepackage{xcolor}
\def\BibTeX{{\rm B\kern-.05em{\sc i\kern-.025em b}\kern-.08em
    T\kern-.1667em\lower.7ex\hbox{E}\kern-.125emX}}
\begin{document}

\title{Improving the Algorithm of Deep Learning with Differential Privacy\\
}

\author{\IEEEauthorblockN{Mehdi Amian}
\IEEEauthorblockA{\textit{INRS-EMT} \\
\textit{University of Quebec}\\
Montreal, Canada \\
mehdi.amian@inrs.ca}}

\maketitle

\begin{abstract}

In this paper, an adjustment to the original differentially private stochastic gradient descent (DPSGD) algorithm for deep learning models is proposed. 
As a matter of motivation, to date, almost no state-of-the-art machine learning algorithm hires the existing privacy protecting components due to otherwise serious compromise in their utility despite the vital necessity. The idea in this study is natural and interpretable, contributing to improve the utility with respect to the state-of-the-art. Another property of the proposed technique is its simplicity which makes it again more natural and also more appropriate for real world and specially commercial applications. The intuition is to trim and balance out wild individual discrepancies for privacy reasons, and at the same time, to preserve relative individual differences for seeking performance. The idea proposed here can also be applied to the recurrent neural networks (RNN) to solve the gradient exploding problem. 
The algorithm is applied to benchmark datasets MNIST and CIFAR-10 for a classification task and the utility measure is calculated.  The results outperformed the original work.
\end{abstract}



\section{Introduction}
In general, privacy might be confused with security as some overlap might be perceived in a general point of view. However, in the computer science and in the context of the data science, these two notions have distinct and different meanings. Basically, security refers to the case where two parties communicate with each other through a communication channel while no third party could reveal information. However, when it comes to privacy, there is no third party. Instead, if a party makes a query and collects more information than those supposed to get, it is referred as breaching the privacy of the respondent party.

Privacy can be discussed in two different ways: the data and the model. As for the data, when a dataset associating with some people is publicly shared, it should be guaranteed that these people are not identifiable through any query over the data. In data privacy domain, an important area is medical records which are publicly shared mostly for research and development purposes. Experience shows that anonymizing the records is not sufficient for preserving the privacy. A famous example is identification of Massachusetts Governor in 1990s by matching a public healthcare dataset (hospital's admission records) with a registry vote database \cite{b1.5}. A naive solution might be to release data with lower resolution e.g. giving a range of age instead of the exact age which is less informative though. This is not effective in practice due to a severe compromise in utility. A crucial challenge in the context of privacy is to establish a reasonable trade-off between privacy and utility i.e. offering an acceptable amount of privacy at the least possible cost of utility.  

On the other hand, privacy on the model ensures that for a publicly shared model which is trained on possibly confidential and sensitive data, no query can disclose the confidential information. As an example, some writing recommendation applications like Gboard (Google mobile keyboard) \cite{b1.6}, are trained on the data (text messages, search flow, emails, etc) of the user as well as other users. Another example is Google's Smart Compose which is a commercial model for email composition recommendation that is trained on emails of millions of users \cite{b1.7}. To enlighten the severity of the problem, think of an adversary attack to reverse the training path and reveal the sensitive data, like credit card number or the social security number, on which the public system is trained. A relatively new emerging area where privacy of model becomes urgent is the federated learning where data inputs come from various users and then are integrated within a federated model. Several studies addressed privacy concerns in federated learning \cite{b5,b6,b7,b8,b9,b4}.

In this study, the spotlight is on the second case on the privacy of the model, and more precisely, deep learning models. The problem that is addressed in the present work is to propose modification for the original DPSGD algorithm to provide a better performance (accuracy) in the deep neural networks. In the proposed approach, the privacy cost (particularly privacy loss which will be denoted by $\epsilon$) theoretically is more than DPSGD but in practice is hard to say, as will be discussed later. DPSGD theoretically provides arbitrary privacy amount but at exhaustive performance compromise which makes it non-applicable to real-world and particularly commercial usages. The accuracy is more desirable when is higher, ideally 1 which is equivalent to a perfect performance, while the privacy cost is expected to be as small as possible, ideally 0 which corresponds to a perfect privacy. Despite the fundamental need, almost no state-of-the-art machine learning algorithms, which are trained on giant datasets containing potentially confidential information, recruits privacy preserving mechanisms because of non-tolerant decline in their performance. For instance, the state-of-the-art language models which are trained on text data and possibly on sensitive data like credit card numbers, email addresses, and physical addresses do not utilize privacy providing techniques \cite{b3.6}. So, for such models, there is a serious real demand for developing privacy components to be deployed in real-world applications.

It is shown that deep learning models are naturally susceptible to memorizing the input data and, as a result, are prone to give in the private data as the response of curious and possibly malignant queries \cite{b3}. This memorization is different from overtraining \cite{b3.5} so it can not be resolved by applying techniques such as regularization. As the data size grows, the vulnerability of the model augments as well. Carlini et al.\cite{b3.6} demonstrated this issue by introducing attack on GPT-2, a language model, released by OpenAI, which is trained on giant public Internet data. They could successfully reveal hundreds of training data including information such as the names, phone numbers, physical address, and email addresses of people. Hence, it seems necessary to introduce privacy preserving units into these networks. Mainly, the privacy is a qualitative and abstract notion. However, when it comes to analysis and modelling, it becomes essential to settle quantitative grounds. There are several quantitative definitions for privacy among which the most popular ones are k-anonymity, l-diversity, t-closeness \cite{b15}, and differential privacy (DP) \cite{b12,b13}. The DP is a well-suited definition of privacy that so far reflects the best the general abstract notion of the privacy in a quantitative manner. Abadi et al \cite{b1} proposed the differential privacy stochastic gradient descent (DPSGD) algorithm which induces the privacy into deep learning models. The algorithm delivers some differential privacy at the cost of some accuracy, of course. The intuition was to clip individual gradients in order to diminish the extreme effect of some individual inputs that might otherwise lead to unintended memorization of the system. It is also further supported by adding noise to sum of the individual gradients. The approach leverages scaling dawn the gradients whose second-order norm is greater than some predefined threshold while keeping other gradients unchanged. The privacy is provided in this way but at the cost of losing some performance. The lost performance raises controversy over the effectiveness of this approach. Some variants are proposed to retrieve the deteriorated accuracy and set up a better trade-off between the performance of the network and the privacy budget \cite{b10,b11}.

Nasr et al \cite{b2} introduced a modified version of the original DPSGD algorithm based on encoding gradient and denoising which restored some lost performance. However, the approach still fails to resolve the issues existing in the original work, plus working at an exorbitant computational cost.  

To provide privacy, the influence of individual inputs should be balanced out in a way that prevents the (unintended) memorization of the system due to extreme unbalanced footprints of some individuals. To preserve the utility, at the same time, it is necessary to retain the relative influence of individuals. These inspire using some smoothing filtering procedures like the tangent hyperbolic function which spreads out the inputs between +1 and -1 no matter how big they are (contributing to privacy) while keeping the superiority order (contributing to utility). 
So, the idea is to incorporate a tangent hyperbolic filtering operation that replaces the gradient clipping step. While looking simple which makes it appropriately fit to real world applications, it offers privacy at a quite modest utility cost. The algorithm is evaluated on MNIST and CIFAR-10 datasets and the results are compared with state-of-the-art.

The paper is organized as following. A brief background for the topic is provided in the next section. Afterwards, the proposed modification is described. Next, the experimental results are presented. The paper, then, is concluded by a discussion over the results and the whole work.
\section{Background}

\subsection{Differential Privacy}

When two datasets are different in only one member, the differential privacy (DP) \cite{b12,b13} guarantees that  the results of queries on them can be arbitrary close. 

\medskip

\textbf{Definition 1.} \cite{b13} A randomized algorithm $\mathcal{A}$ is $(\epsilon, \delta)$ differentially private for any subset $\mathcal{S}$ of its range if for all $\mathcal{D}$ and $\mathcal{D'}$ adjacent datasets in its domain:
\begin{equation}
Pr [\mathcal{A}(\mathcal{D}) \in \mathcal{S}] \leq e^\epsilon Pr [\mathcal{A}(\mathcal{D'}) \in \mathcal{S}] + \delta
\end{equation}

In DP, randomization is essential \cite{b13}. To include the randomness into deep learning algorithm, one strategy is to add noise to gradients. A mechanism with additive zero-mean Gaussian noise is called Gaussian mechanism. 

\medskip

\textbf{Definition 2.} \cite{b13} The $l2-$sensitivity of a function $f$ for any $x$ and $y$ in the domain of the function is defined as follows:

\begin{equation}
    \Delta_2(f) = \max_{x,y, \parallel x - y\parallel_1 = 1} \parallel f(x) - f(y)\parallel_2
\end{equation}

The Gaussian mechanism provides the DP which is defined by the following theorem.
\medskip

\textbf{Theorem} \cite{b13} The Gaussian mechanism with parameter $\sigma$ which is greater than $c\frac{\Delta_2(f)}{\epsilon}$ is $(\epsilon, \delta)$ differentially private, provided that $c^2>2\ln(1.25/\delta)$ for any $\epsilon$ between 0 and 1.

\subsection{ Rényi Differential Privacy}

The privacy metrics in this work are calculated according to Rényi Differential Privacy (RDP) \cite{b14} which is a natural relaxation of differential privacy based on Rényi divergence.

\medskip

\textbf{Definition 3.}  A randomized algorithm $\mathcal{A}$ is $(\alpha, \epsilon)$ RDP with $\alpha \geq 1$, if for any $\mathcal{D}$ and $\mathcal{D'}$ neighboring datasets in its domain, the Rényi divergence satisfies:

\begin{equation}
    \frac{1}{\alpha - 1}\log \mathbb{E}_{\delta \sim \mathcal{A}(\mathcal{D'})}[(\frac{\mathcal{A}(\mathcal{D})}{\mathcal{A}(\mathcal{D'})})^\alpha] \leq \epsilon
\end{equation}

The following proposition links the RDP with DP.

\medskip

\textbf{Proposition} \cite{b14} A mechanism that is $(\alpha, \epsilon)$ RDP is also $(\epsilon+\frac{\log1/\delta}{\alpha - 1}, \delta)$ differentially private for arbitrary value of $\delta$ between 0 and 1.
\section{Contribution}
 The modified DPSGD algorithm is summarized as following. Just as for clarification on terminology used here, the number of microbatches shouldn’t be confused with the microbatch size. The former refers to the number of microbatches within a single batch while the latter denotes the number of samples inside a microbatch.
 
 The algorithm is initialized with (usually) random initial parameters (weights and biases) which are denoted as $\theta$. The dataset is divided into a number of mini-batches. At each trial (whose total number is denoted as $T$), one batch is randomly selected and then is divided into smaller units called micro-batch. Each micro-batch is passed to the network and then the gradient is calculated. To provide privacy, the gradient is passed through the tangent-hyperbloic filter whose role is to balance out and suppress the extreme values of individual gradients while preserving their superiority accordingly. All filtered gradients corresponding to micro-batches within a mini-batch add up to which the noise is added then to ensure the essential randomness which is required to provide the DP. Next step is to update parameters and take a step in the opposite direction of the gradient. The process is recursively continued for all mini-batches. 


\begin{algorithm}[h]
\SetAlgoLined
\SetKwInput{KwInput}{Input}
\SetKwInput{KwOutput}{Output}
\KwInput{parameters $\theta$, learning rate $\eta$, number of microbatches $\mu$, noise standard deviation $\sigma$ }

 initialize $\theta$ randomly\;
 \For{$t\in \{T\}$}{
  $B_t \leftarrow$ take a batch from dataset randomly
  
  $\nabla_{\theta} \leftarrow 0 $
  
  \For{microbatch $b\in B_t$}{
  
  $\nabla _{\theta}^{\mu} \leftarrow$ gradient of microbatch $b$
  
  $\nabla _{\theta}^{\mu} \leftarrow \tanh {\nabla _{\theta}^{\mu}}$
  
  $\nabla_{\theta} \leftarrow \nabla_{\theta} +  \nabla _{\theta}^{\mu}$}
  
  $\nabla_{\theta} \leftarrow \frac{1}{\mu} (\nabla_{\theta} + \mathcal{N} (0,\sigma^2 I)) $
  
  $\theta_{t+1}\leftarrow \theta_{t}-\eta \nabla_{\theta}$
  
 }
 \KwOutput{$\theta_T$ and calculate the total privacy cost $(\delta, \epsilon)$ }

 \caption{Modified DPSGD algorithm}
 \lstset{numbers=left, numberstyle=\tiny, stepnumber=1, numbersep=5pt}

\end{algorithm}

The type of the filter is a key factor in designing the model. As mentioned earlier, to settle a desirable trade-off between privacy and utility, it is essential to adopt an approach to appropriately consider both aspects together. The original DPSGD algorithm is based on restricting the individual influences by forcing the l2-norm to be unity while in the proposed approach the contribution of individuals is balanced out and arranged accordingly by a smooth filtering function. The tangent hyperbolic function (Eq. 4), which is plotted in Fig. 1, is a proper choice as it respects the two sides of the scale at the same time. 

\begin{equation}
    \tanh{x} = \frac{1-e^{-x}}{1+e^{-x}}
\end{equation}

\begin{figure}[H]
\centerline{\includegraphics[width=6cm]{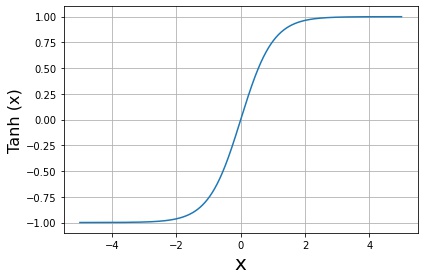}}
\caption{Tangent hyperbolic function}
\label{fig}
\end{figure}

In a privacy protecting perspective, the idea in this work is quite natural and interpretable. The motivation was to develop a privacy preserving structure in deep learning to be applicable in real-world and in particular commercial applications. The naturality of the idea paves the way in other areas than privacy as well. As an important contribution, it can be proposed to be used in recurrent neural networks (RNN) to tackle the gradient exploding problem. By passing the gradients through a TH filter, there remains explicitly no exploding in gradients while preserving the gradient role in converging to the (sub)optimal minimum.

\subsection{Privacy Discussion}

Theoretically, the $l2$-sensitivity 1 condition of DPSGD does not hold when applying the TH filter to gradients. The l2-norm a vector $G$ of size $n$ with entries $g_i$ is defined as following.

\begin{equation}
    \parallel G\parallel_2 = \sqrt{\sum_{i=1}^n g_i^2}
\end{equation}

After passing the gradient element-wise through TH filter, we will have:

$\lvert g_i \rvert < 1 \rightarrow g_i^2 < 1 \rightarrow \sum_{i=1}^n g_i^2 < n \rightarrow  \sqrt{\sum_{i=1}^n g_i^2} < \sqrt{n} $

\medskip

$\rightarrow \parallel G\parallel_2 < \sqrt{n}$

\medskip

This shows that by passing the gradient through the TH function, the $l2$-sensitivity of the gradient can be as much as $\sqrt{n}$ where $n$ is the number of elements of the gradient. As stated earlier, the the $l2$-sensitivity determines how much noise should be added to a Gaussian mechanism for a given amount of privacy. Obviously, as the the $l2$-sensitivity grows, so does the amount of noise to be added. In other words, mathematically, for a given privacy amount ($\epsilon$), the approach in this work requires more noise than DPSGD.  Consequently, more noise will degrade the performance of the system. However, in practice, the reality is different. In fact, the gradient rapidly converges to values close to zero as the training trend approaches the minimum point (local optimum). So, in practice, shortly after starting the training, most gradient elements are placed around and close to zero. In other words, as the training proceeds, the actual norm of gradient will be much lower than the loose bound of $\sqrt{n}$. Hence, in practice, the assumption of unity for the $l2$-sensitivity of the gradient is not far from reality. This fact ensures offering reasonable amount of privacy in practice and improving the performance at the same time which is a remarkable asset for an approach in privacy preserving deep learning area and makes it applicable for real-world and specially commercial applications where the high performance is fundamental and non-negotiable. 





\section{Experimental Results}
\label{sec:figs}
\subsection{Datasets}

In this study two standard datasets, MNIST and CIFAR-10, that are prevalently recognized as benchmark datasets in machine learning tasks. MNIST is a collection of 70000 handwritten digits (0-9) in 10 classes. The training and test sets contain 60000 and 10000 images respectively. The images are in black and white with the size of $28\times28$. CIFAR-10 is a set of 60000 $32\times32$ color images in 10 classes with 6000 images per class. The size of the training and test sets is 50000 and 10000 respectively.

\subsection{Privacy measurement}

In all experiments, the privacy measures are calculated based on Rényi Differential Privacy (RDP). The privacy budget for the proposed approach is calculated in the same way as for the original DPSGD algorithm. In other words, the same amount of noise is considered for DPSGD and the proposed approach. It should be noted that theoretically, the DPSGD assumption of $l2$-sensitivity 1 is not held in the proposed approach. Theoretically, $l2$-sensitivity of a TH filtered gradient is $\sqrt{n}$ where $n$ is the number of the gradient elements. However, in practice, the gradient values are placed quite close to zero such that the bound of $\sqrt{n}$ is too loose for $l2$-sensitivity. In this work, the motivation is to improve the performance of the model in a way that makes it suitable for real-world and commercial applications at a \textit{reasonable} theoretical compromise in privacy. 

The experiments are implemented with TensorFlow on shared GPUs by Google Colaboratory online platform.

A summary of hyperparameter values is listed in Table 1. These are default values in all experiments in this work unless mentioned otherwise. 

\begin{table}[h]

\caption{Hyperparameters list}

\begin{center}
\scriptsize\addtolength{\tabcolsep}{+5pt}

\setlength{\arrayrulewidth}{0.45mm}

{\normalsize
\begin{tabular}{ c c }
 \textbf{Hyperparameter} &\textbf{Value}  \\
\hline
 failure probability $\delta$ & $10^{-5}$  \\ 
 noise covariance $\sigma$& $1.1$ \\ 
 sampling rate for MNIST& 0.427 \% \\
 sampling rate for CIFAR-10& 0.512 \% \\
 minibatch size & 256  \\
 microbatch size & 1  \\

\end{tabular}}
\end{center}
\end{table}


The performance of the proposed algorithm as well the original one on MNIST and CIFAR-10 datasets is shown in Figs. 2 and 3 respectively. The effect of translating inputs into bigger output ranges is also investigated and is plotted for some factors of \textit{c} by which the regular range (-1, 1) is multiplied. As can be seen in these figures, there is no significant effect on accuracy associated with the range of output which is not far from expectation.   


\begin{figure}[h]
\centerline{\includegraphics[width=6cm]{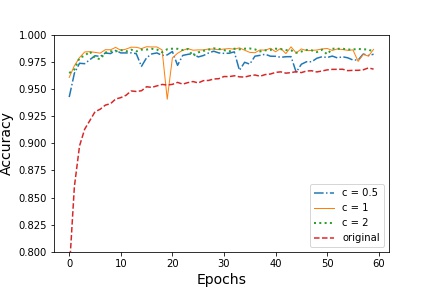}}
\caption{Performance of model with different output ranges of filter compared with original algorithm over training steps for MNIST dataset}
\label{fig}
\end{figure}

\begin{figure}[h]
\centerline{\includegraphics[width=6cm]{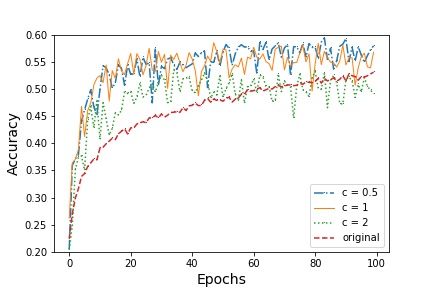}}
\caption{Performance of model with different output ranges of filter compared with original algorithm over training steps for CIFAR-10 dataset}
\label{fig}
\end{figure}

 As the figures show so far, the proposed algorithm offers an improvement in performance compared with the original DPSGD, at the cost of some instability though. Furthermore, a saturation regime in performance results over training epochs is observed such that improving accuracy over more training time seems less likely. As can be seen in Fig. 1, in tangent hyperbolic (TH) function, the outputs almost saturate for the inputs greater than, let's say, 3 (we are not looking for the exact number). In fact, what we call the \textit{active range} (AR) is small in TH. All inputs outside the AR have almost the same output. So, it sounds more natural to enlarge the AR and involve more inputs. The results of enlarging the AR by a factor of \textit{k} are presented in Figs 5 and 6. As these figures imply, expanding the AR contributes to not only stabilizing the results but also retaining the ascending trend which implies achieving even better performance that by taking further training steps. The other observation is that bigger expanding coefficients, on one hand, lead to more stable results, and on the other hand lower the accuracy. So, a new hyperparameter is added here to choose the appropriate expanding coefficient.


\begin{figure}[!h]
\centerline{\includegraphics[width=6cm]{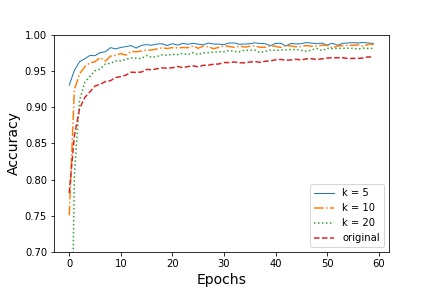}}
\caption{Stabilizing model performance by using different expanding coefficients for MNIST dataset}
\label{fig}
\end{figure}

\begin{figure}[!h]
\centerline{\includegraphics[width=6cm]{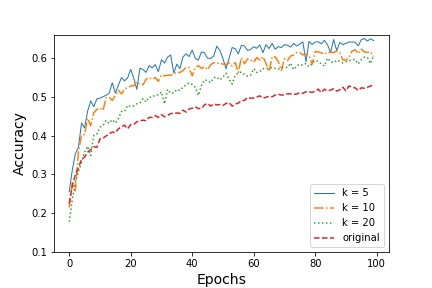}}
\caption{Stabilizing model performance by using different expanding coefficients for CIFAR-10 dataset}
\label{fig}
\end{figure}

\section{Conclusion and discussion}
In this work, a modified version of DPSGD algorithm is presented. The objective is to offer a better performance (higher accuracy) with an acceptable amount of privacy in deep learning models. This study aims at adjusting the existing DP structures to real-world applications which currently lack from a privacy preserving element due to a non-acceptable compromise in performance of such existing approaches. The proposed approach leverages the basic characteristics of privacy and utility together to settle a better compromise between them. The idea is to pass individual gradients through a TH filter to cancel out their odd influences on the system which could otherwise leave a stronger trace on the memory of the system and would raise privacy concerns. While balancing out individual effects, the filter reduces the performance damage by keeping the superiority order of the gradients. Selecting an appropriate filter plays an important role in constructing the model.
Even, the TH in its original shape is not the perfect choice as its AR is relatively small that leads to a fast saturation regime. By enlarging the AR, an improvement in performance in terms of stability is achieved at the cost of some utility loss. Choosing the optimal AR is itself another key decision that needs to be taken at the beginning of the process among other hyperparameters. The results also showed an improvement in the utility measure compared with the original DPSGD algorithm. Theoretically, this approach requires more noise to add to the Gaussian mechanism than DPSGD algorithm. However, in practice the issue is relieved as the $l2$-sensitivity of the gradient is getting well bounded by progressing the training when the gradients are placed around and quite close to zero. As a result, the proposed approach is deemed to better suit the real-world and particularly commercial applications which necessitate high performance. The idea of TH filtering can also be extended to other areas such as RNN to resolve the gradient exploding problem.

Many machine learning and particularly deep learning innovations are discovered through experiments. The proposed approach in this study turns out to work well in practical grounds even though there is a lack of solid theoretical support. The motivation is to contribute to fill the presently existing gap of privacy in real world and specially commercial machine learning models. As future works, some privacy attacks can be designed and applied to deep models equipped with the proposed privacy component in order to observe its efficacy and robustness.

\section{Acknowledgment}

Thank to Martin Abadi, Brendan McMahan, and Nicholas Carlini for inspiration.


\begin{thebibliography}{00}

\bibitem{b1.5}A. Narayanan and V. Shmatikov, `` Robust de-anonymization of large sparse datasets,'' \textit{Proceedings of the 2008 IEEE Symposium on Security and Privacy} pages 111–125, 2008.
\bibitem{b1.6} A. Hard, K. Rao, R. Mathews, S. Ramaswamy, F. Beaufays, S. Augenstein, H. Eichner, C. Kiddon, and . Ramage, `` Federated Learning for Mobile Keyboard Prediction,''  \textit{arXiv:1811.03604}, 2019.
\bibitem{b1.7} M. X. Chen et al, `` Gmail smart compose: Real-time assisted writing,''  \textit{arXiv:1906.00080}, 2019.
\bibitem{b5} K. Bonawitz, F. Salehi, J. Konečný, B. McMahan, and M. Gruteser, `` Federated Learning with Autotuned Communication-Efficient Secure Aggregation,''  \textit{arXiv:1912.00131}, 2019.
\bibitem{b6} Z. Sun, P. Kairouz, A. T. Suresh, and H. B. McMahan, `` Can You Really Backdoor Federated Learning?,''  \textit{arXiv:1911.07963}, 2019.
\bibitem{b7} P. Kairouz et al, `` Advances and Open Problems in Federated Learning,''  \textit{arXiv:1912.04977}, 2019.
\bibitem{b8} S. Augenstein, H. B. McMahan, D. Ramage, S. Ramaswamy, P. Kairouz, M. Chen, R. Mathews, and B. Arcas, `` Generative Models for Effective ML on Private, Decentralized Datasets,''  \textit{ICLR}, 2020.
\bibitem{b9} W. Zhu, P. Kairouz, B. McMahan, H. Sun, and W. Li, `` Federated Heavy Hitters Discovery with Differential Privacy. \textit{arXiv:1902.08534}, 2020.
\bibitem{b4} M. Nasr, R. Shokri, and A. Houmansadr, ``Comprehensive Privacy Analysis of Deep Learning: Passive and Active White-box Inference Attacks against Centralized and Federated Learning,''  \textit{arXiv:1812.00910}, 2020.
\bibitem{b3.6} N. Carlini, F. Tramer, E. Wallace, M. Jagielski, A. Herbert-Voss, K. Lee, A. Roberts, T. Brown, D. Song, U. Erlingsson, A. Oprea, and C. Raffel, `` Extracting Training Data from Large Language Models,'' \textit{ arXiv:2012.07805}, 2020.
\bibitem{b3} N. Carlini, C. Liu, Ú. Erlingsson, and J. Kos, D. Song, `` The Secret Sharer: Evaluating and Testing Unintended Memorization in Neural Networks,'' \textit{Usenix 28}, 2019.
\bibitem{b3.5} I. V. Tetko, D. J. Livingstone, and A. I. Luik, `` Neural network studies. 1. Comparison of overfitting and overtraining,'' \textit{Journal of Chemical Information and Computer Sciences}, vol. 35(5), pages 826–833, 1995.
\bibitem{b15} N. Li, T. Li, and S. Venkatasubramanian, `` t-Closeness: Privacy Beyond k-Anonymity and l-Diversity,'' \textit{IEEE 23rd International Conference on Data Engineering}, pages 106-115, 2007.
\bibitem{b12} C. Dwork, ``Differential privacy,'' \textit{Encyclopedia of Cryptography and Security}, pages 338–340, 2011.
\bibitem{b13}  C. Dwork and A. Roth, `` The algorithmic foundations of differential privacy,'' \textit{Foundations and Trends in Theoretical Computer Science}, 2014.
\bibitem{b1} M. Abadi, A. Chu, I. Goodfellow, H. B. McMahan, I. Mironov, K. Talwar, and L. Zhang, `` Deep Learning with Differential Privacy,'' \textit{Proceedings of the 2016 ACM SIGSAC Conference on Computer and Communications Security (ACM CCS)}, pages 308-318, 2016.
\bibitem{b10} B. Balle, P. Kairouz, B. McMahan, O. Thakkar, and A. Thakurta, `` Privacy Amplification via Random Check-Ins,'' \textit{Preproceeding of the Conference of Advances in Neural Information Processing Systems (NeurIPS 2020)}, 2020.
\bibitem{b11} O. Thakkar, G. Andrew, and H. B. McMahan, `` Differentially Private Learning with Adaptive Clipping,'' \textit{arXiv:1905.03871}, 2019.
\bibitem{b2} M. Nasr, R. Shokri, and A. Houmansadr, `` Improving Deep Learning with Differential Privacy using Gradient Encoding and Denoising,'' \textit{arXiv:2007.11524}, 2020.
\bibitem{b14} I. Mironov, `` Rényi differential privacy. In \textit{2017 IEEE 30th Computer Security Foundations Symposium (CSF)} pages 263–275, 2017.










\end{thebibliography}
\end{document}